\documentclass[11pt]{article}
\usepackage{graphicx,amsmath,amsfonts,amscd,amssymb,bm,cite,epsfig,epsf,url,color,algorithm,algorithmic}
\usepackage{fullpage} \usepackage[small,bf]{caption}
\usepackage{subfigure}
\usepackage{multirow}
\usepackage[mathscr]{eucal}
\renewcommand{\>}{\rangle}

\renewcommand{\c}{\mathcal}
\renewcommand{\b}{\mathbf}

\newcommand{\oT}{\leftarrow}
\newcommand{\R}{\mathbb R}

\newcommand{\x}{\mathbf{x}}
\newcommand{\y}{\mathbf{y}}

\numberwithin{equation}{section}

\def \endprf{\hfill {\vrule height6pt width6pt depth0pt}\medskip}

\title{Sparse \emph{and} silent coding in neural circuits}

\author{Andr\'as L{\H o}rincz$^{1}$, Zsolt Palotai$^{2,3}$ and G\'abor Szirtes$^{1}$\\
  \vspace{-.1cm}\\
  $^1$ E\"otv\"os Lor\'and University, P\'azm\'any P\'eter s\'et\'any 1/C, Budapest, Hungary, H-1117\\
  \vspace{-.3cm}\\
  $^2$ Sparsense Inc., USA\\
 \vspace{-.3cm}\\
  $^3$ ELTE-Soft kft., P\'azm\'any P\'eter s\'et\'any 1/C, Budapest, Hungary, H-1117\\
 }

\date{October 19, 2010}

\begin{document}

\maketitle

\vspace{-0.3in}

\begin{abstract}
Sparse coding algorithms are about finding a linear basis in which signals can be represented by a
small number of active (non-zero) coefficients. Such coding has many applications in science and
engineering and is believed to play an important role in neural information processing. However,
due to the computational complexity of the task, only approximate solutions provide the required
efficiency (in terms of time). As new results show, under particular conditions there exist
efficient solutions by minimizing the magnitude of the coefficients (`$l_1$-norm') instead of
minimizing the size of the active subset of features (`$l_0$-norm'). Straightforward neural
implementation of these solutions is not likely, as they require \emph{a priori} knowledge of the
number of active features. Furthermore, these methods utilize iterative re-evaluation of the
reconstruction error, which in turn implies that final sparse forms (featuring `population
sparseness') can only be reached through the formation of a series of non-sparse representations,
which is in contrast with the overall sparse functioning of the neural systems (`lifetime
sparseness').  In this article we present a novel algorithm which integrates our previous
`$l_0$-norm' model on spike based probabilistic optimization for sparse coding with ideas coming
from novel `$l_1$-norm' solutions.

The resulting algorithm allows neurally plausible implementation and does not require an exactly
defined sparseness level thus it is suitable for representing natural stimuli with a  varying
number of features. We also demonstrate that the combined method significantly extends the domain
where optimal solutions can be found by `$l_1$-norm' based algorithms.
\end{abstract}


\maketitle

\section{Introduction}
\label{s:intro}

To cope with uncertainty in the observations, neural systems are supposed to implement a generative
model \cite{Mumford94Neuronal} for explaining the signals with a set of hypothesis about the world.
The models can then be used to predict future events or spatio-temporal patterns of
observations. Of the many possibilities, the so called sparse coding models seem to get the
strongest support from theory (e.g. \cite{Olshausen02Sparsecodes}) as well as from experiments
(e.g. \cite{Vinje02Naturalstimulation,Smith06Efficient}, but see, e.g., \cite{Berkes09Noevidence}).
Regarding neural activity, sparse coding  refers to a special form of representation where a small
number of neurons are active at a time and the rest are either silent or show weak activity.
\cite{Field94whatisthegoal}. Population codes of this sparse form are suggested to approximate
signals by a linear superposition of the smallest possible subset of features (also called basis
functions) out of a given `dictionary' of features \cite{Olshausen97Sparsecoding}.

Formally, the goal of sparse coding is to represent a signal as a \emph{sparse linear
combination} of  basis elements:

 \begin{equation}\label{e:sparse_app}
min_a \|\b{a}\|_0 \,\,\,\,\,\,\,\,\textrm{subject to} \,\,\,\,\,\,\,\,\b{x}=\bm{\Phi}\b{a}
 \end{equation}
where $\b{x}\in\R^N$ is the signal to be reconstructed, coefficient vector $\b{a}\in\R^M$ is the subject of
sparsification and  $\b{D} \in \R^{N\times M} $ denotes  the so called generative matrix, also referred to as top-down
or reconstruction matrix or dictionary. Columns of matrix $\b{D}$ are \emph{unit norm} basis vectors. Sparsity is
measured by the $l_0$ pseudo-norm (denoted by $\|.\|_0$), which counts the number of nonzero elements in a vector:
$\|\b{a}\|_0 \triangleq \#\{i,\,\textrm{ s.t. } a_i \neq 0\}$.

It has been argued that sparseness is required to decrease the metabolic cost of \emph{spiking},
regardless the statistics of the input signals \cite{Graham06Sparsecoding}.  For most natural
stimuli, the underlying (hidden) statistics of the signals is sparse (see, e.g.
\cite{Olshausen96emergence,Smith06Efficient}), so it is plausible  to seek appropriate sparse
representations for them. One of the most important features of sparse coding is that it can be
used to learn \emph{overcomplete} dictionaries, when the number of features is larger than the
dimension of the stimuli (that is $M>N$). Neural populations are usually highly overcomplete which
may offer many advantages like improved storage capacity or additional decrease of metabolic cost
\cite{Olshausen04Sparsecoding}.

Unfortunately, all these favorable properties come at a cost: finding  optimally sparse
representation of a signal using an overcomplete dictionary is a so-called `NP-hard' combinatorial
problem \cite{Natarajan95Sparseapproximate}. Informally there is no guaranty that any approach
would provide a solution to \emph{all} problems in this form in \emph{polynomial time}.  An
algorithm is said to be solvable in polynomial time if the number of steps required to complete the
algorithm for a given input is $O(d^n)$ for some nonnegative integer $n$, where $d$ is the
dimension of the input.  Polynomial algorithms are considered `fast' in theory, but in practice
there are significant differences among the solutions. For sparse coding, the complexity of the
problem is related to the required sparsity level: $\kappa=K/M$, where $M$ is the size of the
representation, and $K$ is the number of active components. Because exact solutions are difficult
to obtain, different approximate solutions have been proposed, like in
\cite{Olshausen96emergence,girosi98equivalence}. Since sparse coding is very important in many
computation intensive tasks (e.g. bioinformatics, computer vision, data mining), a lot effort has
been put into the development of new methods which can solve at least some problems efficiently.
Recent studies \cite{Donoho06Compressed,Candes06Robust} which can be traced back to
\cite{santosa86linear} have reported a surprising discovery: under certain conditions
sparse signals can be \emph{exactly} reconstructed  via solving the related problem of minimization
in $l_1$-norm (For a comprehensive list on the conditions under which different solutions may
succeed, see \cite{tao_sparse}). In other words the same results can be obtained by minimizing  the
sum of the absolute values of the coefficients  instead of minimizing the number of non-zero
coefficients (minimization in $l_0$ norm). These methods have proven robust in the sense that even
noisy or heavily undersampled signals can be recovered this way. This observation relaxes the
combinatorially hard problem, yet the time complexity of the best solutions is still scales with
the $3^{rd}$ power in the dimension of the input. Since cubic scaling is still
prohibitive for real life applications, several approximate solutions for $l_1$-norm based
sparse coding have been proposed, which may not guarantee to provide the optimal solution, but at
least they are fast as approaching linear complexity.

In addition to the problem of computational complexity, there are other concerns for which most
methods could not be directly implemented in a parallel, neurally plausible manner. As it was
stated in \cite{Rozell08Sparsecodingviathresholding} most sparse coding methods share the following
properties: (1) they include neurally implausible (centralized, non-local) operations, (2) they
fail to produce  exactly zero-valued coefficients in finite time (sub-optimal solutions), (3), for
time-varying stimuli they produce non-smooth variations in the coefficients and (4), they use only
a heuristic approximation to minimizing a desired objective function. In our view, however, there
are two additional concerns that render most sparse coding approaches neurally implausible. First,
efficient methods require a common predefined sparseness level for all inputs. Second, mapping
between input and the sparse representation requires a tremendous amount of synaptic transfers and
for overcomplete representations the sheer amount of bottom-up (that is input$\rightarrow$sparse
representation) calculations may become prohibitive from a metabolic point of view. (dendritic
propagation, excitatory postsynaptic potentials and transmitter release) \cite{Lennie03Thecost} a
neurally plausible sparse coding system should also minimize the required synaptic transfers when
seeking a solution. This requirement essentially implies that during the process of sparsification
the transient activities should also be sparse thus maintaining lifetime sparseness (i.e. sparse
activity of the \emph{individual} neurons over time) \cite{Willmore01Characterising}.

In this study we address these concerns by integrating two approaches for sparse coding. The core
of the proposed solution is based on our earlier method  \cite{Lorincz08spike} which utilizes
probabilistic combinatorial optimization \cite{Rubinstein04Thecrossentropymethod} and explicit
sparsification by minimizing in $l_0$-norm. The other constituent is a variation of fast, but
non-optimal $l_1$-norm based methods whose contribution is an acceleration in feature subset
selection. The resulting method not only addresses the issues articulated by
\cite{Rozell08Sparsecodingviathresholding}, but it also minimizes synaptic transfer between layers
maintaining different representations of the signal and does not require a strict predefined value
for the sparseness level. Our contribution is that we introduce a novel algorithm, which (1)
exploits the favorable properties of $l_1$-norm based solutions, (2) can be implemented in a
neurally plausible (online, parallel and local) fashion and (3) is efficient in terms of
computational time and metabolic cost. The focus is now on forming sparse representations, so we do
not discuss methods to learn the feature set (but see, for example,
\cite{Olshausen97Sparsecoding,Lee07Efficientsparse,Weiss07Learningcompressed,Mairal10Online} on
this matter).

In order to fully expose our ideas, we briefly review in Section~\ref{s:methods}  the main classes
of $l_1$-norm based sparse coding methods. We then shortly review a probabilistic optimization
scheme for combinatorial problems called Cross-Entropy (CE) method
\cite{Rubinstein04Thecrossentropymethod} which can also be applied in sparse coding problems by
explicitly minimizing the number of non-zero coefficients \cite{Lorincz08spike}. This section ends
with the presentation of the integrated solution. In Section~\ref{s:sims} we demonstrate the
superiority of our method over  established methods on artificial, but large-scale problems when
the exact solution is known. In Section~\ref{s:disc}  we discuss  the relevance of our proposal to
distributed computations. We also discuss correspondences between the proposed algorithm and the
underlying computations in sensory processing.  Conclusions are drawn in Section~\ref{s:conc}. In
the Appendix we detail the pseudo-code of the $l_1$-norm based and the CE methods as the main
constituents of our integrated solution.

\section{Methods}\label{s:methods}

In this section we present our notations and describe the so-called $l_1$-norm based solutions to
the problem of sparse coding as defined by Eq.~\ref{e:sparse_app}. In addition, cross entropy
method is presented, which is provably convergent, has been very efficient in probabilistic
combinatorial optimizations and can be used for explicit sparsification by minimizing in
$l_0$-norm.  Finally we present an integrated solution, which is believed to inherit many favorable
properties of the $l_0$ and $l_1$-norm based methods.

\subsection{Notation}\label{s:notations}

Let the $l_p$-norm of vector $\b{a}=(a_1, \ldots , a_M)^T \in \R^M$, where $T$ stands for
transposition, be defined as $\|\b{a}\|_p\triangleq (\sum_{i=1}^M \|a_i\|^p)^{1/p}$.

We use calligraphic letters for index series: \mbox{$\c{J}:=\{(j_1,\ldots,j_J):
j_1,\ldots,j_J\in\{1,\ldots,M\}, J\le M\}$} where the corresponding plain capital letter denotes
the size of the index series and indices are between 0 and $M$.

For any $\mathbb{Z}_+ \ni K(\leq M)$ and index series $\c{K}=(k_1, \ldots , k_K)$ (where $k_n \in
\{1, \ldots , M\}$ and $k_n \neq k_m $ if $n\neq m$) let $\b{a}[\c{K}]$ denote the $\c{K}$-indexed
components of $\b{a}\in \R^M$, i.e., $\b{a}[\c{K}]=(a_{k_1}, \ldots , a_{k_K})^T$. Similarly, for
matrix $\b{D} \in \R^{N\times M}$ and for series $\c{K}=(k_1, \ldots , k_K)$ $(k_n \neq k_m \in
\{1, \ldots , M\})$, let $\b{D}[\c{K}]=(\b{d}_{k_1}, \ldots , \b{d}_{k_K})$.

For a vector $\b{a}\in \R^M$  let us denote the indices of the components arranged in decreasing
order by $\verb"MaxInd"_\c{K}(\b{a})$ , i.e., $\verb"MaxInd"_\c{K}(\b{a})=(k_1, \ldots, k_K)$ with
$a_{k_n}\geq a_{k_m}$ for all $k_n<k_m$ and $n, m \in \{1, \ldots , K\}$. We will refer to this set
as `ordered index series'. Then for matrix $\b{D} \in \R^{N\times M}$, matrix
$\b{D}\left[\verb"MaxInd"_\c{K}(\b{a})\right]$ is in $\R^{N\times K}$ and the columns of this
matrix are ordered according to the values of the first $K$ largest components of vector $\b{a}$.
We will call $\b{D}\left[\verb"MaxInd"_\c{K}(\b{a})\right]$ as `$K$-truncated matrix'.

%To simplify  notations in the following paragraphs, let us introduce a particular index set. Let
%$I^{(*,k)}$ denote the ordered set of indices of the  first $k$ largest coordinates of a vector
%whose coordinates are sorted in descending order. When it is required, an additional superscript
%will denote the iteration number at which the given index set gets updated: $\Lambda^{(*,k,t)}$
%stands for $\Lambda^{(*,k)}$ at iteration $t$.
Pseudo-inverse of matrix $\b{D}$ is defined as $\b{D}^{\dag}=(\b{D}^{T}\b{D})^{-1}\b{D}^{T}$. A
single superscript for matrices and vectors denotes the iteration number: $\b{a}^i$ means vector
$\b{a}$ at iteration $i$. For notational simplicity, iteration number will be dropped and  updates will
be denoted by an arrow pointing to the left ($\oT$), e.g., for update $\b{a}^{i+1}=\b{a}^i +
\b{c}^i$ we use $\b{a} \oT \b{a}+\b{c}$ if no confusion may arise.

%The level set of $\b{a} \in \R^M$ (with parameter $\gamma$) is denoted by  $L_\gamma(\b{a}) := \{ a_m
%\mid a_m \geq \gamma, 1\leq m\leq M \}$.

\subsection{Solution via linear programming} Heuristics (e.g. as in \cite{Olshausen96emergence}) developed for
sparse coding problems are now getting replaced by methods based on the $l_1$-norm substitution
\cite{Rozell08Sparsecodingviathresholding,seeger08bayesian}. In doing so the difficult combinatorial optimization
problem can be recast into a much easier convex optimization task, for which there exist efficient interior
point-type linear programming \cite{Boyd04ConvexOptimization.} techniques. This version of the
optimization reads as
\begin{equation}\label{e:BP}
\min_\mathbf{a} \|\mathbf{a}\|_1 \,\,\,\,\,\,\,\, \textrm{subject to} \,\,\,\,\,\,\,\, \mathbf{x}=\mathbf{D}\mathbf{a}
\end{equation}
or in a relaxed form for the noisy case:
\begin{equation}\label{e:BPDN}
    \min_\b{a} (\|\b{x}-\b{D}\b{a}\|_{2}^{2}+\lambda \|\b{a}\|_1)
\end{equation}
where $\lambda$ is a trade-off parameter controlling the balance between the reconstruction quality
and sparsity. For more details, see, e.g. \cite{Chen01Atomicdecomposition}.

\subsection{Iterative methods} The downside of simple linear programming based solutions is that their remarkable
polynomial computational complexity is cubic at best, which is still prohibitive for large scale
applications. The need for faster decoders --~aiming at linear time operation~-- has brought about
improved linear programming methods (e.g. linear programming with preconditioned conjugated
gradients \cite{Kim07Aninterior-point}) as well as other classes not based on convex programming.
Most importantly, a family of greedy algorithms (e.g.
\cite{Tropp04Greedisgood,Needell09Uniformuncertainty}) --~following the ideas of the Matching
Pursuit (MP) approach \cite{Mallat93Matchingpursuits}~-- became dominant due to their low
complexity and simple geometric interpretation.

The core idea is that reconstruction is built up iteratively: the best element of the dictionary is
selected to minimize the \emph{residual} at iteration $I$: $\b{r}^I=\b{x}-\b{D}[\c{I}] \b{a}^I$,
that is the difference between the signal and its actual estimation made of the combination of the
previously chosen  elements of the dictionary. $\c{I}$ denotes the corresponding index series of
the already selected basis functions, $\b{D}[\c{I}]$ denotes the matrix made of the active basis
functions and thus forming a restricted top-down reconstruction matrix, whereas $\b{a}^I$
represents the corresponding (optimized) multipliers.  Although MP is fast, its approximation
performance can be quite poor. Improved solutions, like Orthogonal Matching Pursuit (OMP)  methods
provide better performance, but usually require longer running times.   In general the iteration of
the different greedy methods consists of the following two steps:

\begin{description}
  \item[1. Select:]  Find  basis vector $\b{d}_b$ that provides the largest projection
  of the residual and whose index is not yet contained in $\c{I}$:
\begin{equation}
 b= \operatorname*{arg\,max}_{j \notin \c{I}}  \|\b{d}_{j}^T
 \b{r}^{I}\|.
\end{equation}
Expand the index set with this new index: $\c{I} \oT \c{I} \cup b$.

  \item[2. Improve basis set and update:]  Expand the sparse basis: \mbox{$\b{D}[\c{I}] \oT \b{D}[\c{I}] \cup \b{d}_{b}$}.
  Compute the new approximation, i.e., the coefficients $\b{a}^{I+1}$ and the
  residual $\b{r}^{I+1}$ by minimizing the approximation error, $\|\b{r}^{I+1}\|_2$ according to the applied method.
\end{description}
Iteration may stop when the allowed number of dictionary elements is reached or when the
approximation error is reduced below a predefined threshold. Note that MP and OMP methods differ in
their update procedure \cite{Mailhe09Alowcomplexityorthogonal}:

\begin{description}
  \item[MP:] $\,\,\,\,\,\b{r} \oT \b{r} - \b{d}_{b} \b{d}_{b}^T \b{r}$,
%  \item[GP] $\b{r}_i= \b{r}_{i-1}-\frac{\\|\bm{\Phi}_{i}^{T}\b{r}_{i-1}\\|_{2}^{2}}{\\|\bm{\Phi}_{i}\bm{\Phi}_{i}^{T}\b{r}_{i-1}\\|_{2}^{2}}\bm{\Phi}_{i}\bm{\Phi}_{i}^{T}\b{r}_{i-1}$
  \item[OMP:] $\b{r} \oT \b{r}-\b{D}[\c{I}] \b{D}[\c{I}]^{\dag} \b{r} \,\,(=\b{x}- \b{D}[\c{I}] \b{D}[\c{I}]^{\dag}\b{x})$.
\end{description}
In MP not only the choice of the feature subset is suboptimal, but also the approximation of the
coefficients. OMP improves upon MP by recomputing all coefficients at every step in order to ensure
orthogonality between the residual and  the columns of $\b{D}[\c{I}]$. While OMP methods achieve
better reconstruction, they impose stronger constraints on the matrices than the original theorem
for $l_1$-norm based optimization does (see, e.g. \cite{Dai09Subspacepursuit}). Further
acceleration can be achieved if more than 1 basis functions are chosen at every iteration as e.g.
in the `stagewise' OMP \cite{Donoho06Sparsesolutionofunderdetermined}. Nevertheless, all of these
iterative procedures are somewhat limited by the following issue. Once a verdict is made, the
chosen features remain in the subset until the algorithms terminates. One mistake thus may tend to
influence the reconstruction quality for many subsequent iteration steps.

%The pseudo-code of MP is given in Fig \ref{f:MP}
%
%\begin{figure}
%  % Requires \usepackage{graphicx}
%  \includegraphics[width=13cm]{MP_pseudocode.eps}\\
%  \caption{The pseudo-code of Matching Pursuit algorithms}\label{f:MP}
%\end{figure}

\subsubsection{Subspace pursuit (SP) methods}\label{sss:SP}

A remedy for the above mentioned problem has been independently proposed in
\cite{Dai09Subspacepursuit} and \cite{Needell08CoSaMP}. These methods assume that at most $K$
components are sufficient to represent the input. The methods enlarge the subset of candidate
features by $K$ \cite{Dai09Subspacepursuit} (or $2K$ \cite{Needell08CoSaMP}) features and then
decrease their number back to $K$ at every iteration. These methods thus work with subspaces. The
method of \cite{Dai09Subspacepursuit} is as follows. First, at iteration $i$
 matrix $\b{D}^{T}$ and residual $\b{r}^{i-1}$ are used to increase the feature
set: features corresponding to the $K$ largest values of $\b{D}^{T}\b{r}^{i-1}$ are retained and
added to the previously selected feature set, $\b{D}^{i-1}$. Second, all the selected $2K$ features
are used for reconstruction using the pseudoinverse and the set is then reduced again  by retaining
only the $K$ largest features. Last, the new residual is computed with the retained features. For
completeness, the corresponding pseudocode  of SP of \cite{Dai09Subspacepursuit}  is provided in
the Appendix (Table~\ref{t:SP_pseudocode}).

%\begin{figure}
%  % Requires \usepackage{graphicx}
%  \includegraphics[width=12cm]{SP_pseudocode3.eps}\\
%  \caption{The pseudocode of the Subspace Pursuit method of \cite{Dai09Subspacepursuit}.}
%  The significant difference between this and other incremental MP methods is the refinement of the already
%  chosen subset of basis by testing the projection of the \emph{original} signal onto the current basis set
%  (Iteration step \#2-3).\label{f:SP_pseudocode}
%\end{figure}

\subsection{Subspace cross-entropy (SCE) method for combinatorial optimization}\label{ss:SCE}

While the different $l_1$-norm based  optimization methods are getting more efficient regarding
their computational time complexity, their neural implementation seems quite limited or even
impossible due to some native shortcomings of the algorithms. On the one hand, linear programming
based   methods employ \emph{centralized and non-local} transformations during the optimization
process (see, in \cite{Rozell08Sparsecodingviathresholding}).

% We see two additional arguments against the biological plausibility of the methods considered.
% First, as we have seen, the improvement of the methods has usually been directed to provide better
% reconstruction quality at a lower computational cost (in terms of input size, $N$). However, the
% computational complexity of most methods is still quite sensitive to the sparseness level, $K$, and
% the relative dimensionality of the overcomplete representation (divergence ratio, $M/N$). On the
% other hand, at the different stages of sensory processing in the brain the divergence ratio can be
% significantly larger compared to what most methods can handle. For example, it has been found
% \cite{Stevens01Anevolutionaryscaling} that the number of neurons in the primary visual cortex (V1)
% increases as the 3/2 power of the number of LGN neurons: the human uses more than about 350 neurons
% in V1 for 1 neuron in the LGN

On the other hand, iterative methods, like MP, OMP, and SP rely on  matrix-vector multiplications,
which are --in theory-- suitable for local rule based implementations. However, in all of these
methods the transposed form of the reconstruction matrix $\b{D}$ is used in all iteration steps.
While reconstruction of the original signal uses only $K\times N$ channels (top-down
connections or active synapses in neural networks) and thus only $K\times N$ multiplications are
required, selecting or updating the coefficients requires all $N\times M$ bottom-up channels to
transmit signals between the layers. Considering the significant metabolic cost of synaptic
activation, the huge $M/K$ ratio may suggest that ecological computations should prefer $K\times N$
top-down reconstructions  over $N\times M$ bottom-up transformations.

There is another issue, too, which is beyond simple linear modeling of the signals. As it was
emphasized in the Introduction, neural systems must be endowed with predictive capacity, for which
probabilistic modeling is needed. In addition, due to the complex interplay between different noise
sources (e.g. exogenous source noise generated by the world and endogenous channel noise generated
by the system itself) neural systems often rely on different `modalities' (parallel signal
channels) and on their internal generative model to `fill in' missing parts of the signals or
correct assumably corrupted ones. In turn, purely deterministic representational models just do not
provide enough information to infer e.g. about the reliability of a given feature.

We thus pursue an  alternative approach which (1) minimizes the number of costly bottom-up signal
transmission and (2) provides probability estimations yet it can efficiently solve the problem of
sparse coding. Our solution combines the subspace pursuit idea with the cross-entropy (CE) method
for combinatorial optimization, which directly tries to minimize the number of active components
($l_0$-norm based optimization).

\subsubsection{Cross-entropy (CE) method}

The CE method is a global optimization technique
\cite{Rubinstein99Thecrossentropymethodforcombinatorial} aiming to find the solution in the
following form:
\[
  { \y}^* := \arg\min_{ \y} f(\y).
\]
where $f$ is a general objective function. The CE method resembles the estimation-of-distribution
evolutionary methods (see e.g. \cite{Muehlenbein98Equation}). As a global optimization method, it
provably converges to the optimal solution
\cite{Rubinstein99Thecrossentropymethodforcombinatorial,Muehlenbein98Equation}.  This method has
been successfully applied in different problems  like optimal buffer allocation,
\cite{Allon05Application}, DNA sequence alignment \cite{Keith02Sequence}, reinforcement learning
\cite{szita07learning} or independent process analysis \cite{szabo06cross}.

While most optimization algorithms maintain a single candidate solution $\y(t)$ at each time step,
the CE method maintains a \emph{distribution} over possible solutions. From this distribution,
solution candidates are drawn at random. By continuous modification of the sampling distribution,
random guess becomes a very efficient optimization method.

One may start by drawing many samples from a fixed distribution $g$ and then selecting the best
sample as an estimation of the optimum. The efficiency of this random guessing depends greatly on
the distribution $g$ from which the samples are drawn. After drawing a moderate number of samples
from distribution $g$, we may not be able to give an acceptable approximation of $\y^*$, but we may
still obtain a \emph{better sampling distribution}. The basic idea of the CE method is that it
selects the best few samples, and modifies $g$ so that it becomes more similar to the empirical
distribution of the selected samples.

For many parameterized distribution families, the parameters of the minimum cross-entropy member
can be computed easily from simple statistics of the elite samples. For completeness we provide the
formulae and the corresponding pseudocode for Bernoulli distributions in the Appendix, as these
will be needed in the sparse feature subset optimization. Derivations as well as a list of other
discrete and continuous distributions with simple update rules can be found in
\cite{Boer04Tutorial}.

\subsection{The SCE method}

Among the iterative methods, SP seems  to provide superior speed, scaling and reconstruction
accuracy  over other methods by directly refining the subset of reconstructing (active) components
at each iteration. Its native shortcomings, though, are the heavy use of the costly, bottom-up
transformation of the residuals at each iteration and the preset number of sparsity $\kappa$.
On the other hand, CE which is less affected by the costly bottom-up transformations, updates the
probability of all active components similarly, regardless their individual contributions to the
actual reconstruction error. In turn, we propose an improvement over both methods which inherits
the flexibility and synaptic efficiency of CE and the superior speed and scaling properties of SP
without their shortcomings. The improved model can be realized by inserting an intermediate control
step in CE to individually update the component probabilities based on their contribution to the
reconstruction error. Hence the explicit refinement of SP is replaced by an implicit modification
through the component probabilities. The pseudocode of the combined method is given in
Table.~\ref{t:sce_pseudocode}.

\begin{table}[h!]
 \hrule \vskip1pt \hrule \vskip1mm
\begin{tabbing}
\= xxx \= xxx \= xxx \= xxx \= xxxxxxxxxxxxxxxxxxxxxx \= \kill
 \> \verb"required:" \\
 \> \> $\mathbf{p} = (p_{1},\ldots,p_{M})$                \>\>\>\> \% initial distribution parameters\\
 \> $\verb"initialize:"$ SP $\verb"and"$ CE                              \\
 \> \verb"for" $t$ \verb" from " $1$ \verb" to " $t_{SP}$                   \>\>\>\>\> \% SP iteration main loop \\
% \>\> input: $\mathbf{p}_{t-1}$  \\
 \>\> \verb"for" $\tau$ \verb" from " 0 \verb" to " $t_{CE}-1$,                   \>\>\>\> \% CE iteration main loop \\
 \>\>\> \verb"execute CE iteration" \\
 \>\> \verb"output: "$\c{K}$                           \>\>\>\> \% CE optimized index set \\
 \>\> $\b{r} \oT \x - \b{D}[\c{K}]\b{D}[\c{K}]^{\dag}\x$ \>\>\>\> \% compute next residual\\
 \>\>  \verb"if " $\| \mathbf{r}^t\|_2 \ge \| \mathbf{r}^{t-1}\|_2 \verb" then quit"$  \>\>\>\> \% check for improvement\\
 \>\> \verb"else: SP-like correction for CE" \\
 \>\>\> $\b{e} \leftarrow \b{D}^T \b{r}$  \>\>\> \% BU step of SP  \\
 \>\>\> $\c{M}=(m_1, .. , m_k, .. , m_M) \oT \c{M}[\b{e}]$             \>\>\>  \% ordered index set of $\b{e}$         \\
 \>\>\> $p'_{m_k} \oT \exp \left(- k/K \right)$ \>\>\> \% auxiliary Bernoulli distribution \\
 \>\>                                                                  \>\>\>\> \hspace{3mm} with $\approx K$ number of 1s on average \\
 \>\>\> $\b{p}' \oT \b{p} + \|\b{r}\|_2 \,\b{p}' $ \>\>\> \% weigh by residual's norm  \\
 \>\>                                                                  \>\>\>\> \hspace{3mm}  to improve distribution\\
 \>\> $\b{p} \oT K \,\b{p}' / \|\b{p}'\|_1$ \>\>\>\>\% normalize for K to draw\\
 \>\>                                                                  \>\>\>\> \hspace{3mm} $K$ number of  1s on average \\
 \> \verb"end loop"
\end{tabbing}
 \hrule \vskip1pt \hrule \vskip1mm
 \caption{Pseudo-code of the subspace cross-entropy (SCE) method for Bernoulli distributions}\label{t:sce_pseudocode}
\end{table}

As the resulting algorithm is not a greedy method we call the algorithm as Subspace Cross Entropy
(SCE) method without the term `Pursuit'. In comparison with the algorithms of the original online
CE method (Fig.~1a in \cite{Lorincz08spike}) and the SP method there are two major differences.
With regard to the online CE method, SCE takes advantage of the SP  idea and updates the Bernoulli
distribution with the help of the reconstruction error: the update of the probability of the basis
functions is proportional to their contribution to the representation of the reconstruction error.
With regard to the SP method, SCE is flexible: it is not restricted to $K$ elements. In order to
keep the $l_0$-norm of the representation around $K$, probability vector $\mathbf{p}_t$ is
normalized to $K$. In every iteration SCE draws random samples from the probability vector. It
might turn out that elite samples have more than $K$ 1s, so SCE is not restricted by this
parameter. Furthermore, one may adjust this parameter since the $\lambda$ trade-off parameter of
cost function \eqref{e:BPDN} controls the balance between sparseness and reconstruction quality.
Such adaptive tuning is not included in the pseudocode (Table~\ref{t:sce_pseudocode}) for the sake
of simplicity

In this setting $K$ becomes a soft, tunable parameter dependent on the actual input and
expectations of the system. In the pseudocode we suggest using an exponential function for
normalization based on the projection of the residual on the components. The exponential function
may be replaced by neural non-linear contrast enhancement algorithms (for an early reference, see,
\cite{carpenter87art2}) like soft winner-take-all methods (WTA) often used for competitive learning
in neural networks in tasks like distributed decision making, pattern recognition or modeling
attention (\cite{Carpenter87Amassively,Riesenhuber99Hierarchical}). WTA networks use mutual
inhibition to select the largest (or for k-WTA the first k largest) elements of the input set. The
sharp nonlinearity of the k-WTA procedure corresponds to the SP expansion method
(Table~\ref{t:SP_pseudocode}).

\section{Simulation results}\label{s:sims}

We performed extensive computer simulations in order to compare the performance of the combined
algorithm to some well established solutions, including direct linear programming methods
(`$l_1$-Magic' \cite{Candes06Robust,Candes06L1Magic}) and SP of \cite{Dai09Subspacepursuit}. To see
the impact of SP on CE, we also included the results of a standard batch CE implementation as we
have already shown the equivalence of the online and batch versions of CE (in terms of global
convergence and accuracy) \cite{Lorincz08spike}. In order to provide unbiased comparison, we tested
these methods on a synthetic problem for which the optimal solution is always known. For each
comparison $10$ reconstruction matrices  were generated. Matrix elements were drawn randomly from
Gaussian distribution with zero average and unit variance then the columns were normalized to 1.
For each matrix $10$ distinct random binary vectors were selected as sparse internal
representations (coefficients). Input signals were then generated as products of the internal
representations and the matrices. Results are averaged over 100 runs.

The dimension of the inputs, the sparse, overcomplete representations, and the number of nonzero
components of the sparse representations ($\dim(\x)$, $\dim(\mathbf{a})$ and $K$, respectively)
influence the computational burden of the test. The task is to reconstruct the signals by
estimating the sparse coefficients. We studied the run time and the reconstruction quality as a
function of overcompleteness and sparseness. All simulations were run on a single core PC. Note
that we did not exploit the parallel nature of the batch CE method, which would add an enormous
speed-up in run time over serial methods. Instead, all algorithms were implemented in Matlab
(Mathworks, Natick, MA) and run in the same environment. Table~\ref{t:params} summarizes the
parameters of the algorithms.

\begin{table}[h!]
\begin{center}
\begin{tabular}{|l|l|l|}
\hline Algorithms & Parameters & Values\\ \hline
\multirow{2}{*}{$l_1$-Magic} & primal dual tolerance  & $0.001$\\
 & max. number of primal-dual iterations  & 50\\
 \hline
\multirow{1}{*}{SP} & parameter free (except $K$) &  - \\ \hline
\multirow{5}{*}{CE} & reconstruction error when CE stops & $\epsilon=0.1/K$ \\
 & elite ratio & $\rho= 0.05$ \\
 & update factor for p & $\alpha=0.1$ \\
 & number of batch samplings & 100 \\
 & sample size in one batch & 500 \\ \hline
\multirow{5}{*}{SCE} & reconstruction error when CE stops  & $\epsilon=0.1/K$ \\
 & elite ratio & $\rho= 0.05$ \\
 & update factor for p & $\alpha=0.9$ \\
 & number of batch samplings & 6 \\
 & sample size in one batch & 100 \\
 \hline
\end{tabular}
\end{center}
\caption{Parameters of the different algorithms. $K$ is the number of nonzero components. The
stopping criteria were either the maximum iteration number or the reconstruction error. We used the
default setting for `$l_1$-Magic' \cite{Candes06L1Magic}. SP was implemented as given in
\cite{Dai09Subspacepursuit}. Parameters of batch CE and SCE methods were optimized for these tests
by hand.}\label{t:params}
\end{table}

%\begin{table}\label{t:params}
%\begin{center}
%  \begin{tabular}{ \| l \| l \| }
%    \hline
%    Algorithms & Parameters \\ \hline
%    L1-Magic & VALAMI \\ \hline
%    SP & VALAMI \\ \hline
%    CE & $\epsilon=0.1/k$, elite ratio $\rho= 0.05$, $\alpha=0.1$, batch size=100. \\ \hline
%SCE & $\epsilon=0.1/k$, elite ratio $\rho= 0.05$, $\alpha=0.9$, batch size=6. \\
%    \hline
%  \end{tabular}
%\end{center}
%\caption{Parameters of the different algorithms. $k$ is the number of nonzero components. The stopping criteria were either}
%\end{table}

\begin{figure}[t!]
  \centering
    \subfigure[][]{\includegraphics[width=60mm]{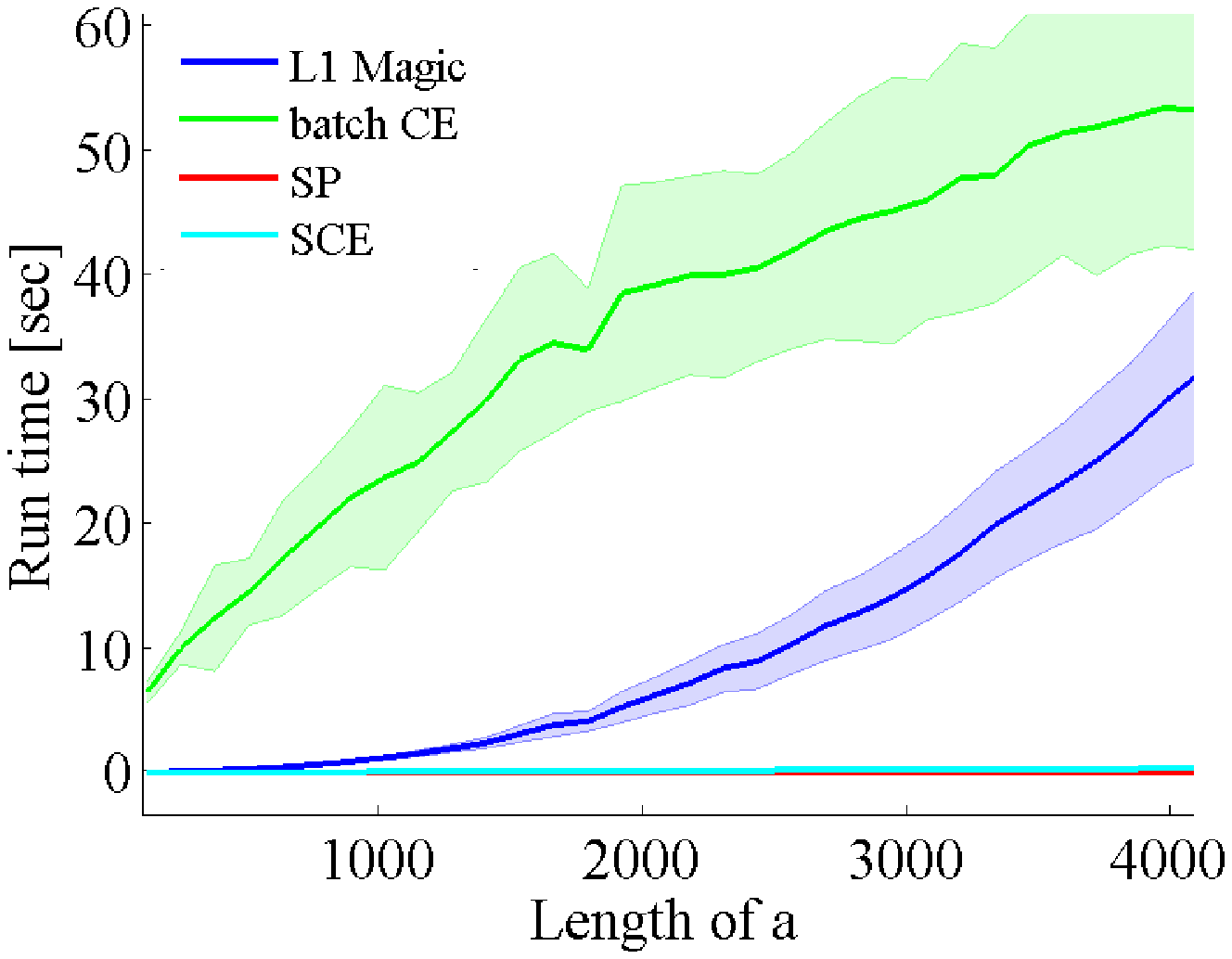}\label{f:runtime_ssize_linear}}
    \subfigure[][]{\includegraphics[width=60mm]{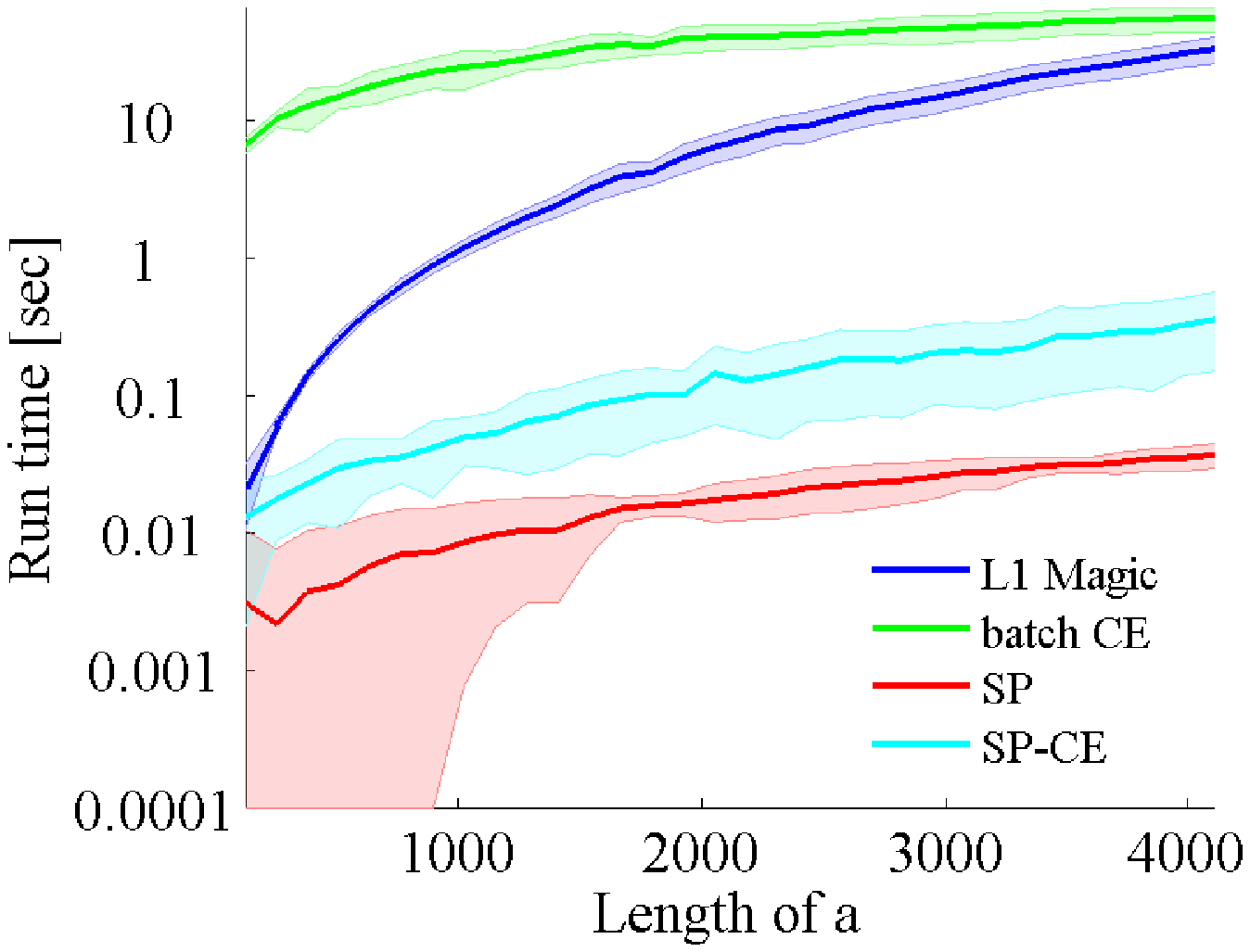}\label{f:runtime_ssize}}
    %\vfill
    %\vspace{0.5mm}
    \subfigure[][]{\includegraphics[width=60mm]{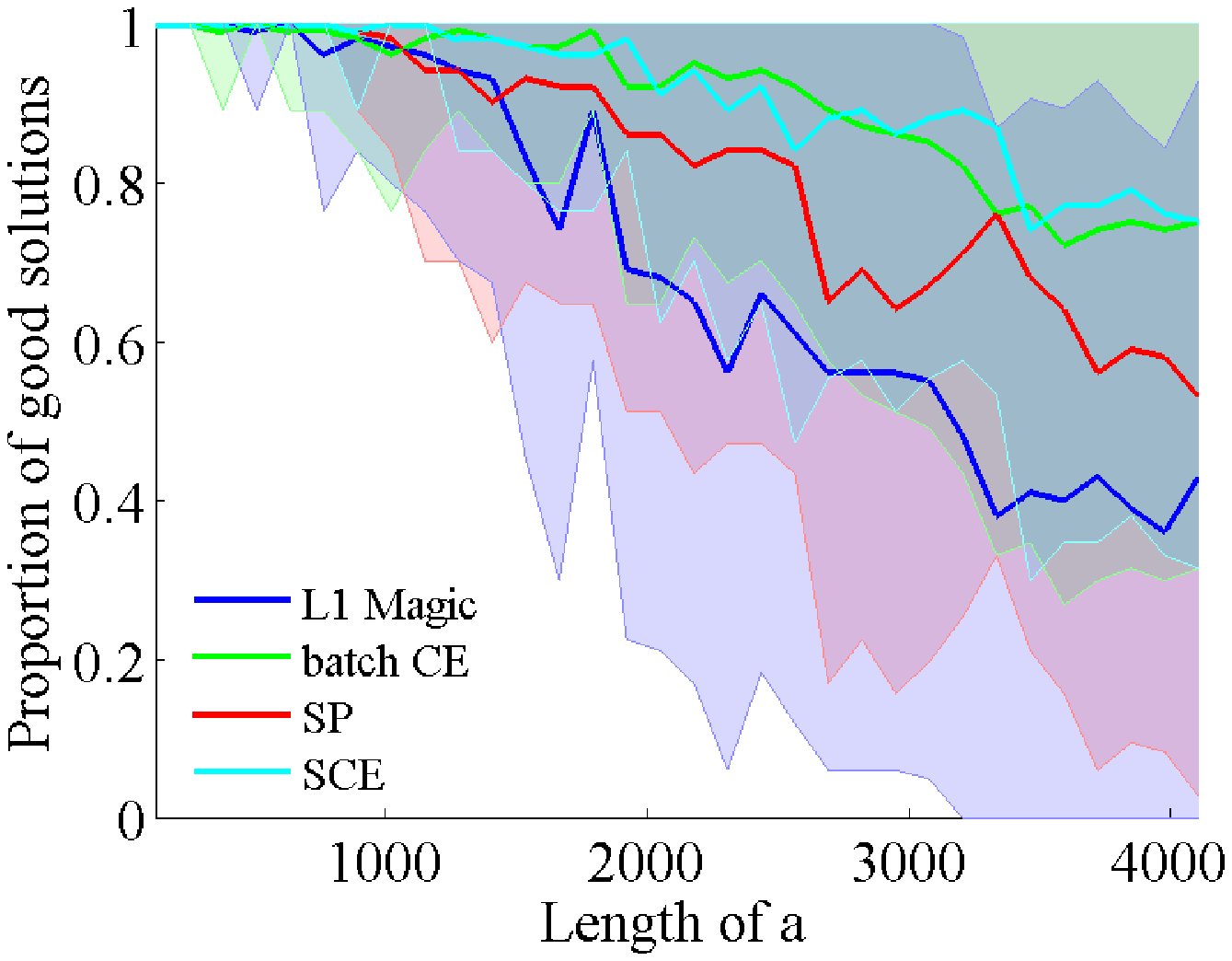}\label{f:goodsolution_ssize}}%\hfill
  \caption{Performance and run time of different sparse coding methods as a function of the size of the representation.
  Compared methods: $l_1$-Magic, (batch) CE, SP and subspace CE (SCE). (a)-(b) Average run time
  for the different methods as a function of overcompleteness, i.e. the size of the internal representation on
  linear and logarithmic plots, respectively.
  Size of input: $\dim(\b{x})=64$, number of active components: $K=8$. (c) Ratio of good solutions (ratio of found good components)
  as a function of overcompleteness. Note: $l_1$-norm and SP methods used the correct number of non-zero components.
  The strict value of this number is not needed for CE and SCE methods}
  \label{f:ssize}
\end{figure}

On Fig.~\ref{f:ssize} three graphs are shown to characterize the performance of $l_1$-Magic, SP, CE
and SCE algorithms. Note that CE should find the optimal solution if allowed to run long enough.
This is not the case for $l_1$-Magic and SP. Conditions of the theorems of $l_1$-Magic are roughly
as follows \cite{Candes06Robust}: solution can be found provided that the dimension of the input is
larger than a constant (on the order of 4) multiplied with the number of non-zero coefficients and
the logarithm of the size of the representation. Note that the condition may be spoiled by
increasing the size of the internal representations. SP, for large representations, may be stuck in
local minima, since it halts if reconstruction error is not improved at a given iteration. Last,
but not least, both $l_1$-Magic and SP assume that the number of non-zero components is given
beforehand, which is not needed for CE and SCE.

The first and second subplots (Fig.~\ref{f:runtime_ssize_linear} and Fig.~\ref{f:runtime_ssize})
show the mean ($\pm$std) run time on linear and logarithmic scales, respectively, as a function of
the size of the internal sparse representation. The logarithmic scale shows the fine structure of
the dependencies, but deviations get distorted. The linear scale demonstrates the order differences
among the methods and also shows the symmetric deviation around the mean.  The third plot
(Fig.~\ref{f:goodsolution_ssize}) depicts the ratio of good solutions as a function of the size of
the internal sparse representation. The different methods fail to provide good solutions if they
reach the maximum iteration number without converging to a solution or converge into an incorrect
solution. Quite remarkably all methods show graceful degradation of performance even for internal
dimensions on the order of 1,000. Nonetheless, there are clear advantages for the CE based methods
for very sparse representations: SP and $l_1$-Magic methods reach `critical sparseness'
\cite{Dai09Subspacepursuit} much earlier. The SCE/CE advantage could be further increased by
allowing longer iterations.

The reconstruction performance of SCE is the same as that of CE within measurement error. However,
SCE runs orders of magnitudes faster than CE and $l_1$-magic. The speed of SCE is due to two
factors. One factor is that SCE is using relatively few time consuming bottom-up transformations.
The other factor is the principled introduction of mutual influence between the probability update
of the individual components that incorporates the advantages of the SP method. In comparison with
pure SP, SCE is only an order of magnitude slower than SP. Nevertheless, we expect to see further
acceleration by extending the algorithm with a predictive module utilizing the CE probabilities not
accessible by the pure SP.

\begin{figure}[h!]
  \centering
    \subfigure[][]{\includegraphics[width=60mm]{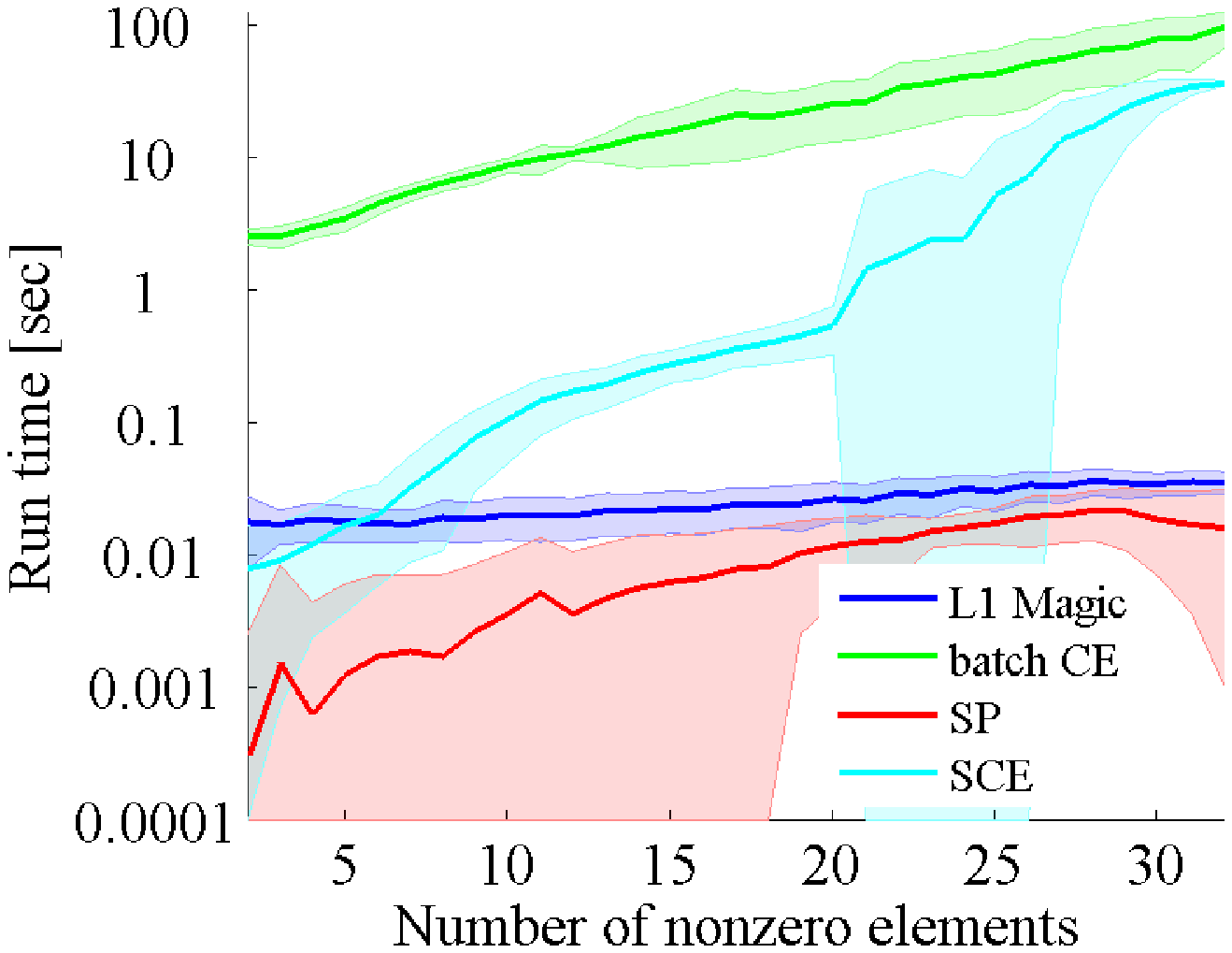}\label{f:runtime_ssize2}}
    %\vfill
    %\vspace{0.5mm}
    \subfigure[][]{\includegraphics[width=60mm]{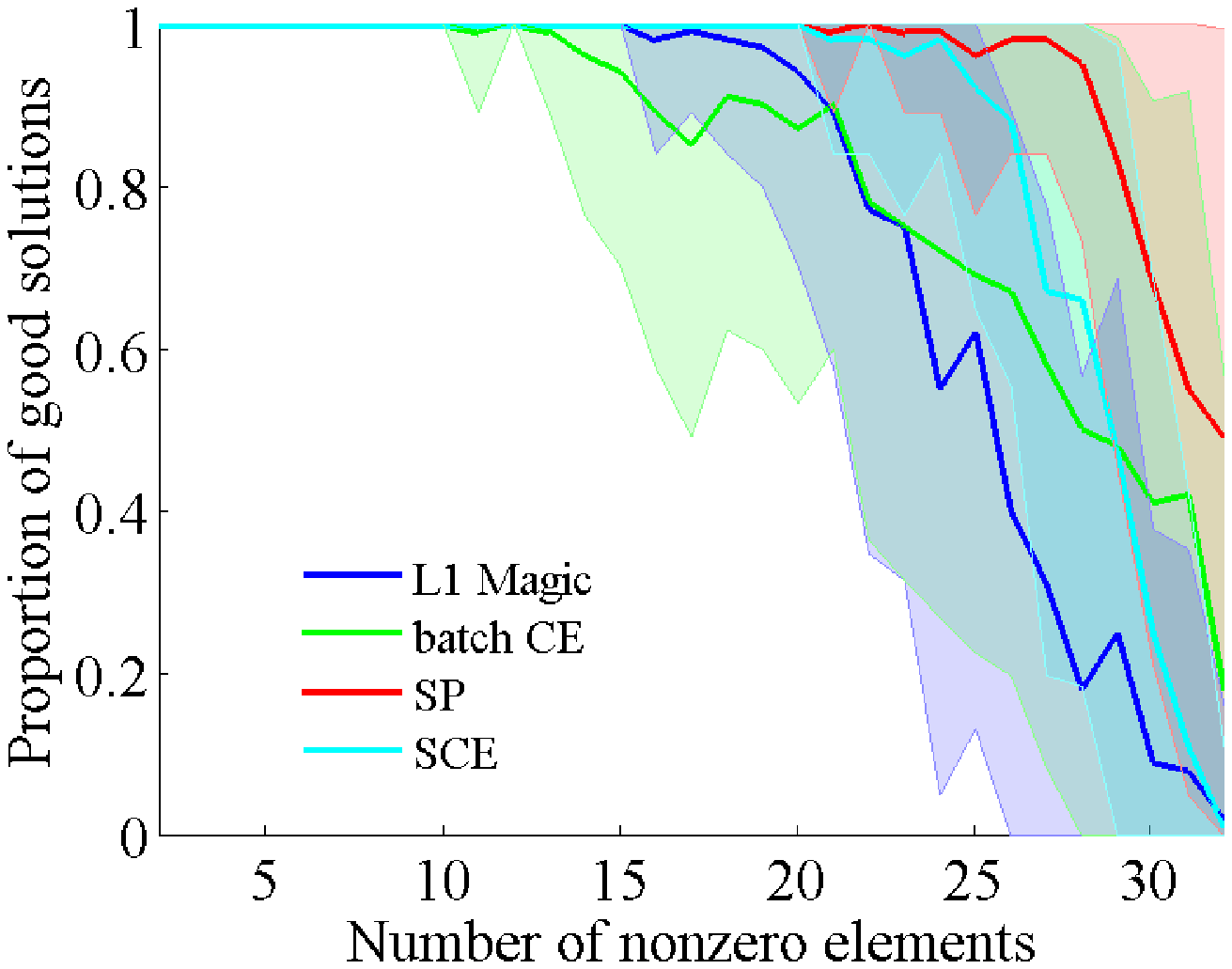}\label{f:goodsolution_ssize2}}%\hfill
  \caption{ Performance and run time of different sparse coding methods as a function of the number of non-zero elements
  of the representation. Compared methods: $l_1$-Magic,
  (batch) CE, SP and subspace CE (SCE). (a) Run time for the different
  methods as a function of sparseness, i.e. the number of non-zero elements of the internal representation.
  Size of input: $\dim(\mathbf{x})=64$, size of the overcomplete representation: $\dim(\mathbf{a})=1024$.
  (b) Ratio of good solutions (reconstruction quality) as a function of overcompleteness. Note: $l_1$-norm and SP
  methods used the correct number of non-zero components. The strict value of this number is not needed for CE and SCE methods}
  \label{f:scomplex}
\end{figure}

The impact of complexity on the different methods has been tested by changing the number of
non-zero components in the overcomplete representations. Figure~\ref{f:scomplex} summarizes the
results. Here again the running time (Fig.~\ref{f:runtime_ssize2}) and the ratio of good solutions
(Fig.~\ref{f:goodsolution_ssize2}) are plotted, but now as a function of the number of non-zero
components. In this test SP seems superior, but the combined method still performs well (increased
number of iterations would allow even better performance at the price of longer run time).
Furthermore as the divergence ratio (size of internal representation/size of input) increases, SP
requires increasingly more iteration for calculating the reconstruction error, thus resulting in an
enormous number of additional synaptic transmissions. This disadvantage does not show up in runtime
using Matlab on a single PC since Matlab is optimized for matrix-vector computations. Let us remark
that performance of SP may quickly deteriorates if the number of non-zero components is not known
beforehand, which is the typical case in real scenarios.

\section{Discussion}\label{s:disc}

According to the expectations brain inspired computations are going to dominate and shape
technology in the near future. While the most compelling features of neural systems are 1) the
extremely low energy cost, 2) massively parallel and distributed organization and 3) outstanding
resilience (robustness against structural and functional perturbations, self-organizing mechanisms
for self-repair, etc), mainstream digital solutions mostly focus on speed, are organized in a
serial fashion and even small scale perturbations can make them malfunction or totally
dysfunctional. In order to make a paradigm shift, we need to have a better understanding of the
structural and functional constituents of the neural systems responsible for the development and
maintenance of these features. In this article we presented a very efficient signal processing
approach designed to form overcomplete, sparse representations. In theoretical neuroscience the
usefulness of such representations has long been recognized (e.g. \cite{Olshausen96emergence}) for
its contribution to increased storage capacity, better pattern separation and higher noise
tolerance.

Recent discoveries in signal processing now suggest another important factor that must be utilized
by evolutionary solutions as well. For most natural signals (i.e. those that are encountered most
frequently by living systems) very efficient \emph{sampling} is possible by exploiting their
intrinsic sparseness. The fact that signals can be recovered by a small set of samples has serious
impact on computational speed as well as on noise tolerance since even partially observed signals
can also be recovered. A few novel models
\cite{Rehn07Anetworkthatuses,Coulter10Adaptivecompressedsensing,Rozell08Sparsecodingviathresholding}
have already started to implement these ideas in neural computations. These studies either deal
with learning efficient dictionaries (in terms of reconstruction quality) or draw parallels between
sensory processing and compressed sensing \cite{Donoho06Compressed,Candes06Robust}. In this article
we argued that efficiency in terms of metabolic cost should also be taken into account when
considering sparse coding mechanisms. Since this constraint is equally important for artificial and
natural systems we wanted to improve $l_1$-norm based solutions by explicitly minimizing the
required number of signal transmission (e.g. by reducing the number of signal projections and
comparisons, especially when bottom-up transmissions are needed). In contrast to $l_1$-norm based
methods, our solution, which is designed to explicitly minimize the active number of components
(i.e., the $l_0$ norm), can be efficiently implemented in a parallel, distributed system.
Furthermore, the CE method ensures convergence to the optimal solution
\cite{Rubinstein99Thecrossentropymethodforcombinatorial,Muehlenbein98Equation}, which may not be
the case for $l_1$-norm based solutions if conditions are not met. What is remained to answer is
whether the SCE algorithm can be implemented in a neurally plausible form.

\subsection{Neural plausibility of the SCE method}\label{s:plausib}

To discuss whether our algorithm makes sense in neural terms, implementation and functional issues
need to be considered. As the implementation is concerned, although a detailed translation of the
model into equations used by computational neuroscience is missing, we conjecture that SCE with
online CE \cite{Lorincz08spike} can be expressed in neuronal form. An important property of the
online variant of CE is that it preserves the convergence properties of the original batch method
\cite{szita08online_TR} although it relies on threshold adaptation and needs to identify elite
samples  and update the probabilities at each iteration. Since thresholds of individual neurons and
neuronal ensembles can be independently and locally adapted and the discrete time updates of the
algorithm can be rephrased in differential forms similar to those describing the changes of
membrane potential we believe that the whole procedure can be translated into neuronal form.

As it was demonstrated our solution is indeed a very efficient sparse coding solution with local
rules supporting distributed implementations. However, SP based update, while considerably
accelerating the computations, does not seem to support lifetime sparseness (at least transiently).
This shortcoming may, for example, be compensated by feedforward inhibitory thresholding
\cite{Rozell08Sparsecodingviathresholding}.

%For example, the probability update of
%Table~\ref{t:CE_method_bernoulli} assumes the form $\frac{dp}{dt}=-p+p'$ if time increment $\Delta
%t$ (which equals 1 in the pseudocode) and parameter $\alpha$ tend to zero with their ratio being
%kept constant.

$k$-WTA is another algorithmic component of our computations. In SCE, its role is  to keep the
number of active components low and to update the probabilities of the components. WTA methods have
been already proposed as a computational primitive in the neural system On a particular use of
$k$-WTA in modeling the hippocampal computations, see \cite{OReilly94Hippocampal}, whereas for a
recent paper on the computational power of neural WTA methods, see, e.g.,
\cite{Maass00Onthecomputationalpower}.

The proposed algorithm uses two distinct quantities: the probability of firing (feature selection)
and the activity that reconstructs the input (coefficients of the features). Whether this dichotomy
can be mapped onto assumed forms of neural computations is an open question. There are many
potential solutions, ranging from local neuronal circuits, through considerations on population
codes \cite{pouget00information,ma06bayesian} to models that neurons themselves represent
multi-layer networks exhibiting a range of linear and non-linear mechanisms
\cite{london05dendritic}. Future work will focus on this issue.

It has been long debated if the nature of neocortical sensory processing is mostly feedforward or
feedback, since even the first spikes after stimulus onset may carry enough information  in e.g.
recognition tasks. \cite{Thorpe97Howcan}. While most arguments are about the expected speed of
information processing, the capacity of the generative networks or the impact of anatomical and
physiological constraints (see, e.g.,
\cite{rao99predictive,koch99predicting,Sillito06always,serre07feedforward,liu09timing}), we find
that a particular combination of feedforward (bottom-up) and feedback (top-down) processing
provides the most favorable solution for sparse coding in terms of \emph{metabolic cost}. Given
that the statistics of natural signals supports sparse representations, ecological solutions, like
this combined process, may have evolutionary advantage over purely feedforward or feedback
solutions.

In respect to the criticisms of \cite{Rozell08Sparsecodingviathresholding}  against previously
proposed neural sparse coding mechanisms our solution 1) is neurobiologically plausible (using
decentralized and local interactions at low synaptic activity), 2) can generate exactly zero-valued
coefficients (since it directly minimizes the number of active components) and 3) it is based on a
principled approach as opposed to heuristic approximations applied in other proposals. However, the
$4^{th}$ issue about non-smooth variations in the coefficients for time-varying stimuli has not yet
been addressed. Its importance stems from the belief that prediction is probably the most
fundamental task of the neural system \cite{Llinas02Iofthevortex}. Hence any model claiming to
explain some aspects of neural representation should provide means to support this central task.
Currently we are seeking methods to make the proposed algorithm applicable for predicting
time-varying sparse signals. Future work will show if the emerging probabilistic representations
can be used e.g. in a belief propagation network thus supporting prediction.

Because all algorithmic steps can be realized by local interactions keeping a low energy profile,
we argue that the emerging architecture might be a good model of neural sparse coding in sensory
information processing. Although the generation of candidate representations and the update of the
component probabilities require recurrent interactions, the constraint to minimize reentrant
activity (i.e. reducing the intra-layer synaptic activity) results in apparent and approximately
correct fast feed-forward working.

Beyond the low metabolic costs of the proposed  algorithm, the $l_0$-norm based optimization
provides  additional flexibility over simple $l_1$-norm optimizations by not fixing the number of
active components beforehand. Methods based on the $l_1$-norm (like SP) may fail if the optimal
sparsity for a given stimuli differs from the predefined value. We note that for all tests
$l_1$-norm based optimization schemes would have provided very poor approximations had they been
restricted to keep a single ratio of non-zero components and the size of the representation
(Fig.~\ref{f:ssize}) or to keep a single number of non-zero components (Fig.~\ref{f:scomplex}).

\section{Conclusions}\label{s:conc}

We have proposed a novel solution for sparse coding by combining our previous work
\cite{Lorincz08spike} on spike based probabilistic optimization using $l_0$-norm with an efficient
$l_1$-norm based method. The motivation behind this integration is that  $l_1$-norm based methods
are computationally attractive and also provide the optimal solution under certain conditions
\cite{Donoho06Compressed,Candes06L1Magic} while our $l_0$-norm based solution using the
probabilistic cross-entropy method provides a potential neural interpretation and does not require
the exact knowledge of the number of non-zero components.

%The probabilistic nature of our solution may offer novel ways to incorporate belief propagation mechanisms by which better initialization of the iteration steps can be achieved thus yielding considerable gain in speed.
The combined method has several outstanding features making it an interesting candidate for
neural sparse coding in the sensory systems: computations are local, distributed and `transmission
efficient' with low metabolic cost. In addition, the algorithm is less affected by constraints of
other alternatives based on $l_1$-norm optimization.

\section{Acknowledgement}

Thanks are due to Zolt\'an Szab\'o for his helpful comments on the manuscript. This research has been partially
supported by the EC NEST `Percept' grant under contract 043261.

\clearpage\newpage

\section{Appendix}\label{s:appendix}

\subsection{Pseudocode for Subspace Pursuit}

\begin{table}[h!]
 \hrule \vskip1pt \hrule \vskip1mm
\begin{tabbing}
 xx \= xx \= xxxxxxxxxxx \= xxxxxxxxxxxxxxxxxxxxxx \= \kill
 \verb"input":\\
 \> $\kappa=K/M$,  $\x \in \mathbf{R}^N$   \>\>\> \% sparsity and signal\\
 \> $t_{SP}$                               \>\>\> \% max iteration number\\
 \> $\b{D} \in \R^{N\times M}$        \>\>\> \% $M$ element dictionary\\
 \verb"initialization":\\
 \> $\c{K} = \verb"MaxInd"_\c{K}(\b{D}^T\b{x})$        \>\>\> \% index set of size $K$\\
 \> $\b{D} = \b{D}[\c{K}]$          \>\>\> \% $K$-truncated matrix\\
 \> $ \mathbf{r} \oT \mathbf{x} - \b{D} \b{D}^{\dag}\b{x}$ \>\>\> \% compute residual\\
 \verb"optimization":\\
 \> \verb"for " $t$ \verb" from " $1$ \verb" to " $t_{SP}$                  \>\>\> \% iteration main loop \\
 \>\> $\verb"compute " \c{K}[\b{D}^T\b{r}]$         \>\> \% index set for expansion\\
 \>\> $\c{L} \oT \c{K} \cup \c{K}[\b{D}^T\b{r}] $     \>\> \% increase set size ($L=2K$)\\
 \>\> $\b{e} \oT  \b{D}[\c{L}]^{\dag} \b{x}$          \>\> \% compute projections\\
 \>\> $\c{K} \oT \verb"MaxInd"_{\c{K}}(\b{e})$        \>\> \% new index set of size $K$\\
 \>\> $\b{D} \oT \b{D}[\c{K}]$                        \>\> \% $K$-truncated matrix\\
 \>\> $ \mathbf{r}t \oT \mathbf{x} - \b{D} \b{D}^{\dag}\b{x}$ \>\> \% compute residual\\
 \>\> $ \verb"if " \mathbf{r}^t=0  \verb" then quit"$     \>\> \% finish is residual is zero\\
 \>\> $\verb"if " \| \mathbf{r}^t\|_2 \ge \| \mathbf{r}^{t-1}\|_2  \verb" then"$  \>\> \% check for improvement\\
 \>\>\> $t=t_{SP}$                                         \> \% no new iteration \\
 \>\>\> $\c{K}^{t_{SP}}=\c{K}^{t-1}$                       \>  \% use previous index set \\
 \>\> $\verb" quit"$                                               \\
 \> \verb"end loop"
\end{tabbing}
 \hrule \vskip1pt \hrule \vskip2mm
  \caption{The pseudocode of the Subspace Pursuit method of \cite{Dai09Subspacepursuit}.
  The significant difference between this and other iterative MP methods is the refinement of the already
  chosen subset of basis by testing the projection of the \emph{original} signal onto the current basis
  set.
 }\label{t:SP_pseudocode}\end{table}

\subsection{CE method for Bernoulli distribution}
Let the domain of optimization be $\b{y}\in\mathscr{Y} = \{0,1\}^M$, and each component be drawn
from independent Bernoulli distributions, i.e., ${\c G} = \textit{BER}\/^{M}$. Each distribution
$g\in \c G$ is parameterized with an $M$-dimensional vector $\mathbf{p} = (p_{1}, \ldots, p_{M})$.
Component $j$ of the sample $\y \in \b{y}\in\mathscr{Y}$ drawn from distribution $g$  thus will be
\[
y_j = \left\{%
\begin{array}{ll}
    1, & \hbox{with probability $p_j$;} \\
    0, & \hbox{with probability $1-p_j$.} \\
\end{array}%
\right.
\]
Having drawn $I$ samples $\y^{(1)}, \ldots, \y^{(I)}$, let
$L_{\gamma}$ denote the level set of the elite samples, i.e.,
\[
    L_{\gamma} := \{ \y^{(i)} \mid  f(\y^{(i)}) \leq \gamma \}
\]
where $\gamma$ denotes a fixed threshold value.
The distribution $g'$ with minimum cross-entropy distance from the
uniform distribution over the elite set has the following parameters \cite{Boer04Tutorial}:
\begin{eqnarray}
    \mathbf{p}' &:=& (p'_1, \ldots, p'_M), \quad\textrm{where} \nonumber\\
    p'_j &:=& \frac{\sum_{\y^{(i)}\in L_{\gamma}} \chi(y^{(i)}_j \!=\! \,\,1)}{\sum_{\y^{(i)}\in L_{\gamma}} 1} =
    \frac{\sum_{\y^{(i)}\in L_{\gamma}} \chi(y^{(i)}_j \!=\! 1)}{\rho\cdot I} \label{e:CEupdate_bernoulli}
\end{eqnarray}
where $\chi(.)$ is an indicator function. In other words, the parameters of $g'$ are simply the
componentwise empirical probabilities of 1s in the elite set. Changing the distribution parameters
from $\mathbf{p}$ to $\mathbf{p}'$ can be too coarse, so in some cases, applying a step-size
parameter $\alpha$ is preferable. The resulting algorithm is summarized in
Table~\ref{t:CE_method_bernoulli}.

\begin{table}[h]
 \hrule \vskip1pt \hrule \vskip1mm
\begin{tabbing}
\= xxx \= xxx \= xxx \= xxxxxxxxxxxxxxxxxxxxxx \= \kill
 \verb"required:" \\
 \> $\b{p} = (p_{1},\ldots,p_{M})$                \>\>\>\> \% initial distribution parameters\\
 \> $T$,                                          \>\>\>\> \%  max iteration number \\
 \> $I$                                           \>\>\>\> \% population size, \\
 \> $\rho$                                        \>\>\>\> \% selection ratio\\
 \> \verb"for " $t$ \verb" from " $1$ \verb" to " $T$                    \>\>\>\> \% CE iteration main loop \\
 \>\> \verb"for " $i$ \verb" from " $1$ \verb" to " $I$\\
 \>\>\> \verb"draw " $\y^{(i)}$ \verb" from " $\textit{BER}^M(\mathbf{p}_t)$      \>\> \% draw $I$ samples \\
 \>\>\> \verb"compute " $f^i := f(\y^{(i)})$                 \>\> \% evaluate them \\
 \>\> \verb"sort "$f^i$\verb"-values" \>\>\> \% order by decreasing magnitude \\
 \>\> $\gamma \oT f_{\rho\cdot I}$         \>\>\> \% level set threshold \\
 \>\> $L_{\gamma} \oT \{\y^{(i)} \mid f(\y^{(i)})\leq\gamma \}$  \>\>\> \% get elite samples \\
 \>\> $p'_j \oT  \bigl(\sum_{\y^{(i)}\in L_{\gamma}} \chi(y^{(i)}_{j} \!=\! \,\, 1)\bigr)/(\rho\cdot I)$ \> \> \> \% get parameters of nearest distrib. \\
 \>\> $p_{j} \oT \alpha\cdot p'_j + (1-\alpha)\cdot p_{j}$ \>\>\> \% update with step-size $\alpha$ \\
 \> \verb"end loop"
\end{tabbing}
 \hrule \vskip1pt \hrule \vskip2mm
 \caption{Pseudo-code of the cross-entropy method for Bernoulli distributions} \label{t:CE_method_bernoulli}
\end{table}

\small
\bibliographystyle{plain}
\bibliography{neural_magic}

\end{document}